\pdfoutput=1

\documentclass[11pt]{article}

\usepackage{naacl2021}

\usepackage{times}
\usepackage{multirow}
\usepackage{latexsym}
\usepackage{graphicx}
\usepackage{todonotes}
\usepackage{booktabs}
\usepackage[T1]{fontenc}

\usepackage[utf8]{inputenc}

\usepackage{microtype}

\title{Abuse is Contextual, What about NLP?\\ The Role of Context in Abusive Language Annotation and Detection}

\author{Stefano Menini \\
  Fondazione Bruno Kessler \\
  Trento, Italy \\
  \texttt{menini@fbk.eu} \\\And
  Alessio Palmero Aprosio \\
  Fondazione Bruno Kessler \\
  Trento, Italy \\
  \texttt{aprosio@fbk.eu} \\\And
  Sara Tonelli \\
  Fondazione Bruno Kessler \\
  Trento, Italy \\
  \texttt{satonelli@fbk.eu} \\}

\begin{document}
\maketitle
\begin{abstract}
The datasets most widely used for abusive language detection contain lists of messages, usually tweets, that have been manually judged as abusive or not by one or more annotators, with the annotation performed at message level.
In this paper, we investigate what happens when the hateful content of a message is judged also based on the context, given that messages are often ambiguous and need to be interpreted in the context of occurrence. 
We first re-annotate part of a widely used dataset for abusive language detection in English in two conditions, i.e. with and without context. Then, we compare the performance of three classification algorithms obtained on these two types of dataset, arguing that a context-aware classification is more challenging but also more similar to a real application scenario.
\end{abstract}

\section{Introduction}
`Abuse is contextual' is one of the key claims reported in  \cite{prabhakaran-etal-2020-online}, where the authors describe the content of a panel with NLP practitioners and human rights experts at the Human Rights Conference RightsCon 2020. A similar remark is made in \cite{jurgens-etal-2019-just} in a recent position paper on current NLP methods to fight online abuse, arguing that NLP research tends to consider a narrow scope of what constitutes abuse, without respecting, for instance, community norms in classification decisions. Similarly, \cite{vidgen-etal-2019-challenges} claim that one of the main research challenges concerning abusive language detection is accounting for context, although they focus mainly on user-level variables and network representations. 

In this work, we present an analysis aimed at better understanding what context is in abusive language detection, and what its effects are on data annotation and classification. We focus on discourse context, i.e. the messages preceding a given tweet, because it has been relatively understudied in abusive language detection, while user-level and network-level features have already been extensively discussed in past works \cite{fehn-unsvag-gamback-2018-effects,del-tredici-etal-2019-shall,10.1145/2567948.2576964,sap-etal-2019-risk,caselli-EtAl:2020:LREC}.

Consider for example the following tweet:

\begin{quote}
\textit{@User1: hell yeah bitches hate that shit}
\end{quote}

Without any context, an annotator may think that it has an abusive content, aimed at offending someone by calling them `bitches', and its offensive intent is strengthened by the use of `that shit'. However, our judgement would change if we knew that the preceding tweet is:

\begin{quote}
\textit{@User2 Ima  stop  calling  females  bitches  next  week sometime. Just tryin to get that bitch word out of my vocabulary}
\end{quote}

Indeed, the context would show that the two users are probably friends and that the message by @User1 is a sort of joke. The main goal of this paper is to analyse what is the impact of cases like this when detecting abusive language on Twitter, both manually and automatically. In particular, the main research questions we want to address are:

\begin{itemize}
    \item[R1:] How would abusive language annotation change if we provided annotators also with the textual context preceding a tweet? Would abusive judgements increase or decrease?
    \item[R2:] What would classification performance be like if we used abusive language datasets annotated with context? Which algorithms would work best and how would the performance change compared to single-tweet classification?
\end{itemize}

To address the above questions, we perform a manual annotation of around 8,000 English tweets in two different conditions, with and without context, and analyse what linguistic phenomena  contribute to different annotations of the same tweets in the two settings (Section \ref{sec:mturk}). We also compare three supervised classification algorithms, described in Section \ref{sec:classifier} and \ref{sec:exper}, and evaluate their different performance on the data (Section \ref{sec:eval1}). Additional remarks on the experiments and a proposal on how abusive language detection tasks should be cast are discussed in Section \ref{sec:discuss}.   
The annotated datasets are made available at \url{https://github.com/dhfbk/twitter-abusive-context-dataset}.

\section{Related work}
The problem of detecting abusive language on social media has recently gained a lot of attention in the NLP community, thanks also to shared tasks on the topic \cite{overviewGermeval,DBLP:conf/evalita/BoscoDPST18,basile-etal-2019-semeval,DBLP:conf/sepln/CarmonaGMEPRS18,zampieri-etal-2019-semeval,zampieri-etal-2020-semeval}  and the consequent availability of benchmarks.  Most of the systems participating in the above tasks rely on deep learning and use embeddings as features, sometimes enriched with  tweet-based surface features like the presence of uppercased words or exclamation marks. For example, OffensEval organisers at Semeval 2019 report that 70\% of participating systems use deep learning \cite{zampieri-etal-2019-semeval}, and that the best performing one in the binary classification task \cite{liu-etal-2019-nuli} is based on fine-tuned
Bidirectional Encoder Representation from Transformer (BERT) \cite{devlin-etal-2019-bert}. Also HateEval organisers at Semeval 2019, focused on hate speech detection against immigrants and women, report that most submitted runs rely on deep learning \cite{basile-etal-2019-semeval}, although the top-ranked system in the binary task for English  uses sentence embeddings as features to be fed into an SVM-based classifier \cite{indurthi-etal-2019-fermi}.

While most approaches to hate speech detection take into account only features related to the single message to be classified \cite{W17-3006,Zhang2018,W18-5104,DBLP:journals/corr/abs-1808-10245}, recent studies have highlighted the importance of context in interpreting the meaning of a message and its possible hateful content \cite{jurgens-etal-2019-just,vidgen-etal-2019-challenges,caselli-EtAl:2020:LREC}. However, context can concern different aspects: some refer to user-level features and profiles \cite{vidgen-etal-2019-challenges}, arguing that an aggressive attitude can be identified at profile level. Some other works suggest that incorporating information about the network structure of a user is beneficial to the detection of hate speech and abusive behaviour \cite{Chatzakou:2017:MBD:3091478.3091487,fehn-unsvag-gamback-2018-effects,del-tredici-etal-2019-shall}. Some other works suggest to consider also the annotator profile, showing that messages written in a specific dialect may be judged differently by annotators with a different ethnicity \cite{sap-etal-2019-risk}, biasing the creation of datasets and therefore of supervised systems. 

Analysing the text surrounding a given message to evaluate its offensiveness, instead, is a rather  understudied problem. Some works have modeled and classified threads from Reddit \cite{Almerekhi:2019:DTT:3342220.3344933} and FoxNews \cite{gao-huang-2017-detecting} to detect words and topics that trigger toxicity, or to find early discussion features that can predict  controversiality, modeling not just the content of the initiating post,
but also the content and structure of the initial responding comments \cite{DBLP:conf/naacl/HesselL19}. These works are however different from ours both in terms of task, of social media platform and of thread structure. Other works have focused on Wikipedia, analysing linguistic cues of ongoing conversations to predict their future trajectory \cite{zhang-etal-2018-conversations}, or trying to assess whether a comment is toxic based only on the preceding messages  \cite{karan-snajder-2019-preemptive}. The most similar work to ours is probably \cite{pavlopoulos-etal-2020-toxicity}, whose goal is to  assess the contribution of context in toxicity detection. Surprisingly, they come to the opposite conclusion, showing that including the textual context increases the perceived toxicity. However, their setting is different, in that they focus on Wikipedia Talk Pages and include as context only the parent comment and the discussion title. Furthermore, the corpus that they create with a setting similar to ours, i.e. by crowd-annotating with and without context, contains only 250 comments. The same authors acknowledge that larger annotated datasets would be needed to estimate more accurately how often context amplifies or mitigates toxicity.

\section{Data Extraction and Analysis}\label{sec:data}

In order to analyse the role of context on data annotation and classification,  we focus on the dataset presented in \cite{founta2018large}, which originally  included around 100K
tweets annotated with four labels: hateful, abusive, spam or none. The authors used a bootstrapping approach to sampling tweets, which were labelled by several crowdsource workers and then validated. 
Differently from other hate speech datasets, this was not created starting from a set of predefined offensive terms or hashtags so as to reduce bias, a main issue in hate speech datasets  \cite{wiegand-etal-2019-detection}. Also for this reason, the dataset has been widely used in experiments for hate speech classification in English.

Since the dataset includes the tweet IDs, we used this information to query the Twitter API and retrieve the complete thread including the given message, so to leverage also the context of each tweet. We then mapped the hateful and abusive tweets to our abusive class, while removing the spam instances. An overview of the dataset is reported in Table \ref{tab:statistics}. The API was queried between August and September 2019 and the retrieval was problematic because some of the tweets were not available anymore, confirming similar remarks in previous works \cite{klubickaexamining}.

 \begin{table}[t]
 \centering
 \small 
 \label{table:1}
 \begin{tabular}{lrrr} 
 \hline \hline
  & Original  & Retrieved & At least 1  \\
 &  Dataset &  & before   \\
 \hline \hline

Abusive & 31,985 & 12,357 & 1,931 \\
Not abusive & 67,814 & 37,091 & 6,087 \\
Total & 99,800 & 49,448 & 8,018 \\
Duplicates & 2,017 &  &  \\

 \hline \hline
 \end{tabular}
  \caption{Statistics on three versions of the dataset by  \cite{founta2018large}: original, retrieved by us, subset having a previous context} 
\label{tab:statistics}
 \end{table}

After removing the tweets no longer available online and the ones without a preceding context, we obtained a new dataset consisting of 8,018 annotated tweets.
As expected, the removal of messages affects more the abusive category, for which only 38\% of the tweets could be retrieved, while for the majority non abusive class 55\% of the messages could  be found. This may depend on the fact that users decided to remove messages from their account, but also that some profiles were suspended in compliance with Twitter policy. 

We also observed that  the original dataset contains duplicated tweets (2,017 in total), probably because both original and re-tweeted messages were annotated, and we therefore removed them.

\section{Manual (Re)annotation}
\label{sec:mturk}
Since one of the goals of this work is to understand whether the textual context plays any role in detecting abusive messages online, we perform a manual re-annotation of the tweets with context. In particular, during the task, annotators not only read the single tweet but also the preceding context and are asked to judge the abusiveness of the message taking the whole conversation into account.
We use Amazon Mechanical Turk platform to perform the annotation of 8,018 tweets with context. 

The guidelines provided to crowd-workers explicitly require the tweet to be annotated as abusive or not considering also what was written before in the thread.
To ensure a good quality of collected annotations, we introduced gold standard examples in the task (one every five tweets) and restricted the access to English native speakers. All requirements introduced by Mechanical Turk for tasks containing adult content, for example adding a warning in the task title, were implemented. For each tweet we collected 3 annotations, and the majority label was finally assigned. Inter-annotator agreement was 0.713 multi-rater free-marginal kappa with 85.6\% of labels in agreement. 

In order to enable a fair comparison between annotation with and without context, we re-annotated the same tweets using the same  methodology without providing crowd-workers with the tweet context. This second annotation was performed selecting the same pool of crowd-workers as for the first task, so to reduce variability due to annotators' bias and preferences. The task was launched around three months after the completion of the first one, making it unlikely that crowd-workers remember the labels assigned to the same tweets with context. This re-annotation without context was performed also to make sure that the two annotated datasets were tagged following exactly the same guidelines. Indeed, the original annotation provided in \cite{founta2018large} was performed by adjusting incrementally the definition of the abusive categories to be annotated. Although they could be mapped to a binary classification, the way in which the annotation process was structured may lead to different results compared to our setting.

\begin{table*}[h]
    \centering
    \small 
    \begin{tabular}{lcrrrcrrr}
    \hline
    \hline
    
    & \multicolumn{4}{c}{Annotations without context (SINGLE-TWEET)} & \multicolumn{4}{c}{Annotations with context (CONTEXT)} \\ 
    
    \cmidrule(lr){2-5} \cmidrule(lr){6-9}
          & At least 1 & \multicolumn{3}{c}{\# previous tweets:} &At least 1 & \multicolumn{3}{c}{\# previous tweets:} \\
&  tweet before & 1 & 2-5 & 6+    & tweet before & 1 & 2-5 & 6+   \\
\cmidrule(lr){2-5}
\cmidrule(lr){6-9}
Abusive & 1,472 & 699 & 608 & 165 & 800 & 404 & 324 & 72 \\
Not Abusive & 6,546 & 2,794 & 2,747 & 1,005 & 7,218 & 3,089 & 3,031 & 1,098\\
Total & 8,018 & 3,493 & 3,355 & 1,170 & 8,018 & 3,493 & 3,355 & 1,170\\
\hline \hline
\end{tabular}
\caption{Comparison of re-annotated tweets providing crowd-workers with no context (left) or preceding context (right)}
\label{tab:dati_riannotati}
\end{table*}

In Table \ref{tab:dati_riannotati} we report the distribution of abusive and non-abusive messages in the dataset annotated with and without context (CONTEXT and SINGLE-TWEET respectively). 
When annotated with context, the abusive class contains 672 tweets less than in the annotation without context, and its weight decreases from 18\% to 10\%. This shows that the main effect of context is to reduce the perceived offensiveness of messages, enabling annotators to understand the tone of the conversation and sometimes the relation between users. The example reported below shows clearly this phenomenon: if we consider the last tweet (in italics) in isolation, we may think that @User2 is aggressively terminating a phone call. Instead, looking at the context we notice that this tweet is just a funny reply to the call mentioned by @User1.\\

\begin{quote}
\noindent \textit{@User1:} My next call: `Hi. I sent an email. I got no response. Is that because you're fucking handling it, or fucking ignoring it?' Wish me luck.\\
\textit{@User2:} \textit{`I'm fucking hanging up now.' $\ast$click$\ast$}
\end{quote}

\normalsize{Another example, reported below, shows how the ironic tweet by @User2 may be labeled as abusive without the previous context:}
\\

\begin{quote}
\noindent \textit{@User1:} Penn State trustee says he's not `totally out' of sympathy for Sandusky victims.\\
\textit{@User2:} \textit{Yeah. Those damn victims...they're the worst! $\ast$shakes fist in the air$\ast$}
 \end{quote}

Interestingly, the longer the context, the more tweets are considered non abusive by annotators: for the group of tweets having a context longer than 5 tweets, only 6\% of them are annotated as abusive. This percentage increases up to 10\% when the context is between 2 and 5 tweets, and 12\% when only one tweet is in the context.

Other cases include tweets whose meaning was unclear or underspecified, for example because of the presence of anaphoric expressions, ellipsis, incomplete messages split across several tweets, etc. In all these cases, the presence of context allows for a more informed decision, reducing the interpretation effort required to annotators. A general trend that we observe is that, whenever a profanity is present in a tweet, it is more likely to be annotated as abusive in the SINGLE-TWEET setting, while it is often labeled as non abusive in the CONTEXT configuration. 
To measure this, we compute the mean square contingency coefficient (Phi coefficient) aimed at capturing the association between the presence of profanities and the `abusive' label. We download the Google profanity words list\footnote{\url{https://github.com/RobertJGabriel/Google-profanity-words/blob/master/list.txt}} and check for each tweet if a profanity is present or not. The Phi coefficient is 0.77 for the SINGLE-TWEET setting, showing a very strong positive correlation between profanities and abusive labels, while it is only moderate (0.39) in the CONTEXT configuration.  This finding is in line with  past works, showing that the offensiveness of swearing is context-dependent \cite{pamungkas-basile-patti:2020:LREC} and that  profanities are frequently used among people with a strong social relationship without any offensive goal \cite{bak-self}, but rather to signal an informal attitude among the members of a community or even intimacy \cite{pei-jurgens-2020-quantifying}.

\section{Classification algorithms}\label{sec:classifier}

As a next step in our study, we want to analyse the classification performance obtained with different algorithms on the newly annotated datasets: SINGLE-TWEET  was annotated by providing crowd-workers with tweets in isolation, and the same information will be used by the classifiers, similar to the standard setting of hate speech detection competitions \cite{zampieri-etal-2019-semeval,zampieri-etal-2020-semeval}. The second, CONTEXT, was annotated by showing crowd-workers the tweet to be classified as well as all the preceding tweets retrieved from Twitter. Likewise, the classifiers will be fed also with the context at training and test time.

We compare two widely used deep-learning approaches, i.e. BERT \cite{devlin-etal-2019-bert} and Bidirectional LSTM (BiLSTM) \cite{graves2005framewise}, as well as a non-neural one based on Support Vector Machines \cite{10.1023/A:1022627411411}. Details on the different algorithms are reported below.

\par \textbf{BERT}: This is a language representation model developed by Google Research \cite{devlin-etal-2019-bert}, whose deep learning  architecture obtained state-of-the-art results in several natural language processing tasks including sentiment analysis, natural language inference, textual entailment \cite{devlin-etal-2019-bert} and hate speech detection \cite{liu-etal-2019-nuli}. BERT can be fine-tuned and adapted to specific tasks by adding just one additional output layer to the neural network. We use the base model of BERT for English,\footnote{Uncased, 12-layer, 768-hidden, 12-heads, 110M parameters.} trained on 3.3 billion words, which is made available on the project website.\footnote{\url{https://github.com/google-research/bert}} For the CONTEXT data, we adopt the configuration proposed in \cite{devlin-etal-2019-bert} for sentence pair classification tasks: the input tweets are given to the classifier as a pair of text chunks, with the first text of the pair containing the context preceding the tweet, and the second one containing the target tweet. For SINGLE-TWEET data, only the string of the tweet is given in input to BERT.

\par \textbf{Bidirectional LSTMs}:  BiLSTMs \cite{graves2005framewise} are made of a combination of two Long Short-Term Memory networks, a particular case of recurrent neural network, i.e. a classifier optimized to process sequences. Using this combination, BiLSTM connects information both from previous and following statuses simultaneously. 
For CONTEXT data, we test BiLSTM in two configurations.
In the \textit{first setting}, we use a sequence of word vectors as input. Regardless the span of the context considered, the entire conversation (the preceding context plus the tweet) is considered as a single text, and each word is represented in the network through word embeddings. The network is configured with three bidirectional layers with respectively 128, 64 and 32 neurons.

In the \textit{second setting} the context and the tweets are represented again as a sequence of word embeddings. Instead of considering them as a single text, we keep the tweet and its preceding messages separated. Both the context and the tweet move through two bidirectional layers of respectively 64 and 32 neurons, before being merged into one to generate a prediction. Both configurations are tested with different values of dropout and with different optimizers (see details in the following Section). For SINGLE-TWEET data, only the first setting is adopted.

\par \textbf{SVM}: For the classification we use LIBSVM \cite{chang2011libsvm}.  As for the CONTEXT data, the tweet and its context are represented by concatenating a  300-dimension sentence embeddings generated from the context and a 300-dimension sentence embeddings for the tweet 
(see Section \ref{sec:exper} for more details). For SINGLE-TWEET, the classifier is fed only with a vector of 300 dimensions.

In our experiments, we also evaluated other approaches, i.e. Feed Forward Networks, Long Short-Term Memory networks and Gated Recurrent Units. All performed worse compared to the classification algorithms described above, therefore we do not include them in the paper.

\section{Experiments}
\label{sec:exper}

\subsection{Preprocessing}

Before running the classification experiments, we preprocess the tweets as follows: we split hashtags using Ekphrasis \cite{gimpel2010part}, which recognises the tokens in a hashtag based on Google n-grams.
The same tool is also used to replace mentions with \textit{``username''} and urls with \textit{``url''}. We also replace every instance of \textit{``money''}, \textit{``time''}, \textit{``date''} (and in general any \textit{``number''}) with a generic label.
We convert each emoji into a textual description using the Python emoji package,\footnote{\url{https://github.com/carpedm20/emoji/}} as suggested in previous work \cite{singh-etal-2019-incorporating}.

Then, since \textbf{BERT} classifier accepts as input sentences or sets of sentences, we just pass the pre-processed tweet text or its context as is.
For the two other approaches, the text is converted into vectors in two ways, both using FastText, a library for efficient learning of word and sentence representations \cite{bojanowski2017enriching, grave2018learning}. In particular, all vectors are extracted starting from the pre-trained embeddings obtained from the Common Crawl corpus.\footnote{\url{https://fasttext.cc/docs/en/crawl-vectors.html}} Since \textbf{SVM} takes in input sentence embeddings, we convert the context and the current tweet into sentence embeddings of 300 dimensions each. Since \textbf{BiLSTM} takes in input word embeddings, each word is represented as a vector of size 300 and a tweet (or context) is then obtained  by merging the resulting set of vectors into a matrix, used as input for the machine learning algorithm.
In the \textit{first setting}, a single matrix is generated, containing all the vectors extracted from both the tweet and the context (if any). In the \textit{second setting}, the matrices of the tweet and the context are generated separately, two different neural networks are created and then merged in the last (common) layer.

\subsection{Experimental Setup}
\label{subsec:contr}

The dataset we use contains 8,018 tweets, that in one of the two configurations are preceded by a context. Note that the class balance of SINGLE-TWEET and CONTEXT is different (details in Table \ref{tab:dati_riannotati}).
As for the CONTEXT data, we experiment with different context lengths, ranging from 1 (i.e. only the preceding tweet) to all. Since not all contexts have the same length, we replace missing tweets in the context with empty strings when a configuration requires a context length that is higher than the available one. 
This simulates a real scenario, in which not every tweet has a context of the same size.

All the configurations and algorithms are evaluated randomly dividing the data into training (80\%), validation (10\%) and test (10\%). Each experiment is repeated three times with different random splits, and the results are averaged.

For BiLSTM,\footnote{The BiLSTM is created with Keras 2.2.5,  and runs with Tensorflow-gpu 1.10.1. } different experiments with a dropout ranging from 0 to 0.5 have been compared. We report only the best results, obtained with dropout = 0.3. The loss function is binary cross-entropy and two optimizers (i.e. SGD and Adam) have been evaluated, although we report only the results obtained with SGD which yields a better performance. The BiLSTM is tested with up to 20 epochs, the results in Table \ref{tab:nocontext} and \ref{tab:t2} are obtained with 10 epochs.
For BERT the best performance, reported in Table \ref{tab:nocontext} and  \ref{tab:t2}, is obtained with 3 epochs (we tested up to 10 epochs), a batch size of 16, a max length of 256 and learning rate 2e-5. Every component of the code relying on randomization uses 42 as seed.
For SVM, we test different types of kernels (i.e. linear, polynomial or radial) and cost parameter (up to 100), reporting only the results obtained with the best configuration, i.e. linear kernel and $C=50$.\footnote{For the experiments we use 4 Tesla P40. The BiLSTM (single text) takes, according to the context size, around one minute per epoch with 2.5M - 4.9M total parameters. The BiLSTM (separate context) takes, according to the context, from 5 to 18 minutes per epoch using 15.1M - 21.5M total parameters. BERT was able to run in 5 to 10 minutes per epoch.}

\section{Evaluation}\label{sec:eval1}
We report in Table \ref{tab:nocontext} the classification results obtained on the SINGLE-TWEET dataset, while in Table~\ref{tab:t2} the classifier performance on the CONTEXT dataset is presented, with a context window ranging from one tweet to all those available in the preceding context.

\begin{table*}[h]
\def\arraystretch{1.2}

\centering
\scalebox{0.80 }{

\begin{tabular}{c|ccc|ccc|ccc}
\hline \hline
Classifier & P\_noAbuse & R\_noAbuse & F1\_noAbuse & P\_Abuse & R\_Abuse & F1\_Abuse & P & R & F1\\
\hline 
\textbf{BERT} & 0.971 & 0.952 & 0.961 & 0.791 & 0.865 & 0.826 & 0.881 & 0.908 & \textbf{0.893} \small{$\pm$ 0.007}\\
\textbf{BiLSTM} & 0.952 & 0.958 & 0.955 & 0.796 & 0.775 & 0.785 & 0.874 & 0.866 & 0.870 \small{$\pm$ 0.004} \\

\textbf{SVM}  &	0.943	&	0.970	&	0.956	&	0.836	&	0.721	&	0.773	&	0.889	&	0.845	&	0.864	\small{$\pm$	0,013}\\
\textbf{Majority} & 0.816 & 1.000 & 0.899 & 0.000& 0.000& 0.000& 0.408 & 0.500 & 0.449\\
\hline \hline
\end{tabular}
}\caption{Classification results on SINGLE-TWEET dataset, average of three runs (with $\pm$ StDev). Precision, Recall and F1 are macro-averaged.} \label{tab:nocontext}
\end{table*}

\begin{table*}[h]

\def\arraystretch{1.2}

\centering
\scalebox{0.8}{

\begin{tabular}{c|ccc|ccc|ccc}
\hline \hline 

Context & P\_noAbuse & R\_noAbuse & F1\_noAbuse & P\_Abuse & R\_Abuse & F1\_Abuse & P & R & F1\\
\hline \hline
\multicolumn{10}{c}{\textbf{BERT}} \\
\hline

1 & 0.951 & 0.970 & 0.960 & 0.643 & 0.517 & 0.572 & 0.797 & 0.743 & \textbf{0.766} \small{$\pm$ 0.017}\\
2 & 0.952 & 0.965 & 0.958 & 0.610 & 0.528 & 0.565 & 0.781 & 0.747 & 0.761 \small{$\pm$ 0.017}\\
3  & 0.950 & 0.967 & 0.959 & 0.620 & 0.511 & 0.560 & 0.785 & 0.739 & 0.759 \small{$\pm$ 0.025}\\
All  & 0.950 & 0.967 & 0.958 & 0.619 & 0.508 & 0.554 & 0.784 & 0.738 & 0.756 \small{$\pm$ 0.018}\\

\hline 
\multicolumn{10}{c}{\textbf{SVM}} \\
\hline

1 &	0.926	&	0.964	&	0.945	&	0.436	&	0.265	&	0.329	&	0.681	&	0.614	&	0.637	\small{$\pm$ 0.019} \\
2 &	0.928	&	0.962	&	0.945	&	0.446	&	0.291	&	0.352	&	0.687	&	0.627	&	0.649	\small{$\pm$ 0.024} \\ 
3 &	0.928	&	0.966	&	0.946	&	0.466	&	0.287	&	0.355	&	0.697	&	0.626	&	0.651	\small{$\pm$ 0.024} \\
All &	0.928	&	0.966	&	0.947	&	0.475	&	0.291	&	0.360	&	0.702	&	0.629	&	0.654	\small{$\pm$ 0.018} \\

\hline
\multicolumn{10}{c}{\textbf{BiLSTM (single text)}} \\ \hline

1  & 0.939 & 0.956 & 0.947 & 0.494 & 0.404 & 0.439 & 0.716 & 0.680 & 0.693 \small{$\pm$ 0.039} \\
2  & 0.938 & 0.957 & 0.948 & 0.488 & 0.397 & 0.433 & 0.713 & 0.677 & 0.690 \small{$\pm$ 0.045} \\
3  & 0.939 & 0.958 & 0.948 & 0.499 & 0.403 & 0.441 & 0.719 & 0.680 & 0.695 \small{$\pm$ 0.040} \\
All  & 0.937 & 0.960 & 0.948 & 0.500 & 0.381 & 0.428 & 0.718 & 0.671 & 0.688 \small{$\pm$ 0.046} \\

\hline 
\multicolumn{10}{c}{\textbf{BiLSTM (separate context)}} \\ \hline

1  & 0.929 & 0.974 & 0.951 & 0.581 & 0.282 & 0.355 & 0.755 & 0.628 & 0.653 \small{$\pm$ 0.040} \\
2  & 0.931 & 0.974 & 0.952 & 0.615 & 0.302 & 0.378 & 0.773 & 0.638 & 0.665 \small{$\pm$ 0.039} \\
3  & 0.926 & 0.974 & 0.950 & 0.548 & 0.254 & 0.314 & 0.737 & 0.614 & 0.632 \small{$\pm$ 0.064} \\
All  & 0.928 & 0.971 & 0.949 & 0.549 & 0.274 & 0.333 & 0.738 & 0.623 & 0.641 \small{$\pm$ 0.050} \\

\hline
\multicolumn{10}{c}{\textbf{Baseline (majority class)}} \\ 
\hline
& 0.900 & 1.000 & 0.947 & 0.000 &0.000&0.000&0.450&0.500& 0.474\\

\hline \hline
\end{tabular}
} \caption{Classification results on CONTEXT  dataset, average of three runs (with $\pm$ StDev). Precision, Recall and F1 are macro-averaged. `Context' column shows the number of preceding tweets considered as context.}
\label{tab:t2}
\end{table*}

Some considerations concerning the classifiers' performance hold for both datasets. Indeed,  BERT-based classification yields the best score both in the SINGLE-TWEET and in the CONTEXT setting, while SVM achieves the worst performance. For SVM, we tested also a different way to model the context, creating a single 300-dimensional FastText embedding for the context and the tweet, but the classifier performance decreased.    
Concerning BiLSTM, modelling the context and the tweet to classify as a  single text is better than separating them in two layers. 
In both settings, BERT yields a much better F1 on the abusive class, the minority one, showing that it can deal with unbalanced classes better than the two other approaches. We tested other strategies to give the tweet and the context in input to BERT, for example by merging them in a single text rather than split in two columns, following the setting proposed for single sentence classification tasks in \cite{devlin-etal-2019-bert}, but the overall performance decreased.

Concerning the differences between the two settings, we observe that with SINGLE-TWEET all classifiers achieve a better performance than with CONTEXT, and this result is consistent across all algorithms and configurations. Our results suggest that the classification based only on single messages may be easier than the context-aware one, not only because of a simpler setup of the classification algorithms, but also because all the lexical information necessary to interpret the abusiveness of a tweet are present in the same message. This makes it easier for the classifier to capture the connections between lexical elements and the corresponding class, for example between profanities and the (supposed) abusiveness.  Indeed, when including only single tweets, F1 is 0.893, in line with the results reported by participants at the last SemEval task on abusive language detection for English, where 60 systems achieve F1 $>$ 0.90 \cite{zampieri-etal-2020-semeval}. Using a context-aware annotation and classification, instead, leads to a drop in performance, affecting in particular the abusive class. 

We observe also that the context size does not show a consistent impact. Furthermore, different context lengths  do not have a statistically significant effect, which we measure for each classification algorithm using approximate randomization test.

Since in the CONTEXT dataset the number of abusive tweets is only 800, the lower performance achieved compared with SINGLE-TWEET may be due to the greater class imbalance and the lower number of abusive instances in the training set, rather than to differences in the task. To measure this effect, we run a further comparison using the same BERT configuration on both datasets, but downsampling them so that they contain the same number of abusive and non abusive instances, i.e. 800 abusive, of which 633 for training, and 6,456 non abusive, of which 5,232 for training. The results of this comparison are reported in Table \ref{tab:balanced} as the average over three runs. This additional comparison confirms that on the SINGLE-TWEET dataset BERT yields a much better performance. It also shows that this difference is inherently due to the way in which the task is modeled, rather than to the size and balance of the classes. Indeed, the performance on this smaller dataset is even slightly better than on the one obtained on both datasets before downsampling.

\begin{table*}[h]
\def\arraystretch{1.2}

\centering
\scalebox{0.79 }{

\begin{tabular}{c|ccc|ccc|ccc}
\hline \hline
\multicolumn{10}{c}{\textbf{SINGLE-TWEET} dataset} \\ \hline
Context Size & P\_noAbuse & R\_noAbuse & F1\_noAbuse & P\_Abuse & R\_Abuse & F1\_Abuse & P & R & F1\\ \hline
0  & 0.963 & 0.971 & 0.967 & 0.859 & 0.822 & 0.839 & 0.911 & 0.897 & 0.903 \small{$\pm$ 0.006}\\
\hline \hline
\multicolumn{10}{c}{\textbf{CONTEXT} dataset} \\ \hline
Context Size & P\_noAbuse & R\_noAbuse & F1\_noAbuse & P\_Abuse & R\_Abuse & F1\_Abuse & P & R & F1\\
\hline 
 1  & 0.956 & 0.958 & 0.957 & 0.591 & 0.585 & 0.587 & 0.774 & 0.771 & 0.772 \small{$\pm$ 0.001}\\
 2  & 0.961 & 0.954 & 0.957 & 0.590 & 0.629 & 0.608 & 0.775 & 0.792 & 0.783 \small{$\pm$ 0.013}\\
 3  & 0.957 & 0.960 & 0.958 & 0.598 & 0.582 & 0.589 & 0.777 & 0.771 & 0.774 \small{$\pm$ 0.026}\\
 All  & 0.954 & 0.961 & 0.957 & 0.598 & 0.555 & 0.575 & 0.776 & 0.758 & 0.766 \small{$\pm$ 0.014}\\

\hline \hline
\end{tabular}
}\caption{Classification results using BERT on downsampled SINGLE-TWEET and CONTEXT   datasets}\label{tab:balanced}
\end{table*}

\section{Discussion}
\label{sec:discuss}

The manual re-annotation of the dataset by \cite{founta2018large} including the preceding tweets highlighted the relevant role played by context when deciding whether a message is abusive or not. However, additional work is needed to find better approaches for effectively including context in the classification. Previous analyses on Wikipedia \cite{karan-snajder-2019-preemptive,pavlopoulos-etal-2020-toxicity} highlighted similar difficulties in modelling the textual content of a thread for automated classification. Our case is even more challenging because of a major difference between Twitter and other platforms: while on other social networks, whose goal is to explicitly discuss a topic, it is rather easy to extract a conversation graph, identifying posts and related replies (see e.g. the conversation trees from Reddit discussions in \newcite{DBLP:conf/naacl/HesselL19}, the structure of Twitter threads is less stable. Indeed, the list of tweets that we extract to build a context is not necessarily a sequence of replies in a coherent conversation, but could include gaps created by deleted messages. This could make the relation between an abusive message and its context not clear, since the next messages are not necessarily a reaction to what was said before.

Reshaping abusive language annotation and detection as a task that should take the context into account is in our view relevant also because it would contribute to mitigating the problem of \textit{disembodiment} in NLP \cite{waseem2020disembodied}, i.e. the fact that both data and machine learning encode subjective content that is however presented as universalised and objective, with the potential to create social marginalisation. For the abusive language case, considering tweets in isolation would oversimplify the way in which online abuses happen in reality, ignoring the fact that context gives a more informed perspective on the circumstances that led to a supposed abuse. Although we are aware that abusive language annotation is affected by different types of biases, ranging from data sampling \cite{wiegand-etal-2019-detection,ousidhoum-etal-2020-comparative} to annotators' preferences \cite{waseem:2016:NLPandCSS}, we believe that including the local context in the task definition could be a first step towards a fairer approach to data annotation and classification.

\section{Conclusions}\label{sec:conclusions}
In this work we investigate the role of textual context in abusive language detection on Twitter. We first manually re-annotate the tweets in the dataset from \cite{founta2018large} for which the preceding tweets could be retrieved online. Annotation was performed by the same crowd-workers in two different time periods, asking them to take into account the context and then to annotate the tweets in isolation. A comparison between the two datasets shows that the context is sometimes necessary to understand  the real intent of the user, and that it is more likely to make a tweet non abusive, even if it contains profanities. We therefore recommend that, in future annotation efforts for abusive language detection, the textual context of messages is also provided to annotators. This would have an impact also on the implementation of classification systems, which should foresee strategies to embed contextual information: the task would be more challenging, but also more similar to a real application scenario. Our classification experiments suggest also that the high performance achieved with BERT-based systems in `standard' abusive language detection may be overly optimistic, because when context is given in input to the same classifiers and they are evaluated on context-aware datasets, their performance drops dramatically.     

In the future, we hope that the release of the datasets presented in this paper and our initial experiments will foster research in the direction of context-aware abusive language detection, including both the creation of guidelines and annotated data, as well as novel classification approaches. 

As a next step, we plan to investigate novel ways to model context, for example by analysing the role of direct responses containing an explicit mention of the user. We also plan to further extend this approach to other platforms, to assess whether platform-specific differences can be observed regarding the role of context, along the line of what has already been observed for single messages \cite{DBLP:conf/clic-it/CorazzaMCTV19}.

\bibliography{naacl2021}
\bibliographystyle{acl_natbib}




\end{document}